\documentclass[10pt,twocolumn,letterpaper]{article}

\usepackage{wacv}
\usepackage{times}
\usepackage{epsfig}
\usepackage{graphicx}
\usepackage{amsmath}
\usepackage{amssymb}

\wacvfinalcopy 

\def\wacvPaperID{479} 

\usepackage{fixltx2e}
\usepackage{stmaryrd}
\usepackage[pagebackref=true,breaklinks=true,letterpaper=true,colorlinks,bookmarks=false]{hyperref}
\newcommand{\method}{{ODGI}\xspace}
\newcommand{\yolo}{YOLO\xspace}

\newcommand{\sdd}{\textsc{sdd}\xspace}
\newcommand{\vedai}{\textsc{vedai}\xspace}
\newcommand{\iou}{\texttt{IoU}}
\newcommand{\map}[1]{\texttt{MAP#1}}

\newcommand{\comment}[1]{}
\newcommand{\ott}{\method-tt\xspace}
\newcommand{\oyt}{\method-yt\xspace}
\newcommand{\oms}{\method-35-35\xspace}
\newcommand{\omb}{\method-100-35\xspace}
\newcommand{\mobilenet}{mobile}
\usepackage{xcolor}
\definecolor{links}{rgb}{0.16, 0.36, 0.69}
\definecolor{citations}{rgb}{0.9,0.3,0.2}
\hypersetup{colorlinks, linkcolor=links, pdfborder={0 0 0}, citecolor=citations}
\usepackage{makecell}
\usepackage{multirow}
\usepackage{tikz}

\newcommand\copyrighttext{%
\scriptsize Accepted to be Published in: Proceedings of the 2020 IEEE Winter Conference on Applications of Computer Vision, March 2-5, 2020, Snowmass Village, CO, USA
 
2020 IEEE. Personal use of this material is permitted. Permission from IEEE must be obtained for all other uses, in any current or future media, including reprinting/republishing this material for advertising or promotional purposes, creating new collective works, for resale or redistribution to servers or lists, or reuse of any copyrighted component of this work in other works.”}
\newcommand\copyrightnotice{%
\begin{tikzpicture}[remember picture,overlay]
\node[anchor=south west, xshift=45pt, yshift=24pt] at (current page.south west) {\fbox{\parbox{\dimexpr\textwidth-\fboxsep-\fboxrule\relax}{\copyrighttext}}};
\end{tikzpicture}%
}

\ifwacvfinal\fi
\begin{document}

\title{Localizing Grouped Instances for Efficient Detection in Low-Resource Scenarios}

\author{
  Am\'{e}lie Royer\\
  IST Austria\\
  {\tt\small aroyer@ist.ac.at}
  \and
  Christoph H. Lampert\\
  IST Austria\\
  {\tt\small chl@ist.ac.at}
}

\maketitle
\ifwacvfinal\thispagestyle{empty}\fi

\begin{abstract}
   State-of-the-art detection systems are generally 
   evaluated on their ability to \textbf{exhaustively} retrieve objects
   \textbf{densely} distributed in the image, across a wide variety of 
   appearances and semantic categories.
   Orthogonal to this, many real-life object detection applications, for example in
   remote sensing, instead require dealing with large images that contain only
   a few small objects of a single class, scattered \textbf{heterogeneously}
   across the space.
   In addition, they are often subject to strict \textbf{computational
   constraints}, such as limited battery capacity and computing power.

  To tackle these more practical scenarios, we propose a novel flexible detection scheme
    that efficiently adapts to variable object sizes and densities: 
  We rely on a sequence of detection stages, each of which
  has the ability to predict \textbf{groups of objects as well as individuals}.
  Similar to a detection cascade, this multi-stage architecture spares
  computational effort by discarding large irrelevant regions of the image 
  early during the detection process.
  The ability to group objects provides further
  computational and memory savings, as it allows working with lower
  image resolutions in early stages, where groups are more easily
  detected than individuals, as they are more salient.
  We report experimental results on two aerial image datasets, and show that
  the proposed method is as accurate yet computationally more efficient
  than standard single-shot detectors, consistently across three different
  backbone architectures.
\end{abstract}
\copyrightnotice

\section{Introduction}
\label{sec:intro}
As a core component of natural scene understanding,
object detection in natural images has made remarkable progress in 
recent years through the adoption of deep convolutional networks. 
A driving force in this growth was the rise of large public benchmarks,
such as PASCAL VOC~\cite{pascal} and MS COCO~\cite{coco}, 
which provide extensive bounding box annotations for objects 
in natural images across a large diversity of semantic categories 
and appearances. %

However, many real-life detection problems exhibit drastically different 
data distributions and computational requirements, for which 
state-of-the-art detection systems are not well suited,
as summarized in \hyperref[fig:intro]{Figure \ref{fig:intro}}.  
For example, object detection in aerial or satellite imagery 
often requires localizing objects of a \emph{single class}, 
\eg, cars~\cite{cars},  houses~\cite{houses} or swimming
pools~\cite{pools}. 
Similarly, in biomedical applications, only some specific objects
are relevant, \eg certain types of cells~\cite{cancer}.
Moreover, input images in practical detection tasks are often of much 
higher resolution, yet  contain small and sparsely distributed objects of
interest, such that only a very limited fraction of pixels is actually relevant,
while most academic benchmarks often contain more salient objects and cluttered scenes. 
Last but not least, detection speed is often 
at least as important as detection accuracy for practical applications. 
This is particularly apparent when models are meant to run 
on embedded devices, such as autonomous drones, which have 
limited computational resources and battery capacity. 
\newlength{\imgheight}
\setlength{\imgheight}{1.8cm}
  \def \trimbot {2.2cm}
\begin{figure}[t]
  \resizebox{\linewidth}{!}{%
  \begin{tabular}{|c||p{3cm}|p{3cm}|p{3cm}|}
    \hline
    \textit{Dataset} & \makecell{\textbf{VEDAI}~\cite{vedai}\\\includegraphics[height=\imgheight, trim={0 \trimbot{} 0 0}, clip]{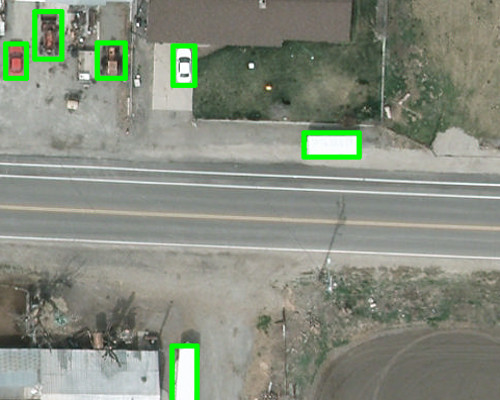}} & \makecell{\textbf{SDD}~\cite{stanford}\\\includegraphics[height=\imgheight, trim={0 \trimbot{} 0 0}, clip]{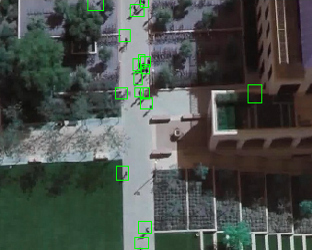}} & \makecell{\textbf{MS COCO}~\cite{coco}\\\includegraphics[height=\imgheight, trim={0 \trimbot{} 0 0}, clip]{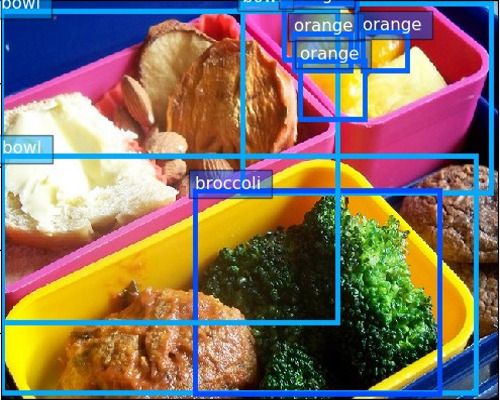}}\\
    \hline
    \hline
    \makecell{\textit{Avg. object size}\\\small\textit{(fraction of image pixels)}} & \centering small (0.113\%) & \makecell{small (0.159\%)} & \makecell{large (14.96\%)} \\
    \hline
    \makecell{\textit{Avg. empty cell ratio}\\\small\textit{(on a 16x16 uniform grid)}} & \centering 95.1\%& \centering 97.1\% & \makecell{49.4\%}\\
    \hline
    \hline
    \textit{Object distribution} & \multicolumn{2}{c|}{\makecell{Few, specific, classes. Sparse and\\ heterogeneous object distribution}} & \makecell{high object density\\ across many and\\ varied categories}\\
    \hline
    \makecell{\textit{Typical }\\\textit{applications}}& \multicolumn{2}{c|}{\makecell{remote sensing, traffic monitoring,\\agriculture, medical imaging}} & \makecell{robotics, artificial\\intelligence, HCO} \\
    \hline
    \makecell{\textit{Resource}\\\textit{constraints}}& \multicolumn{2}{c|}{\makecell{imposed by hardware, \\ \eg limited battery, no GPU}} & \makecell{generally none}  \\
    \hline
  \end{tabular}}
\caption{Recent benchmarks and challenges highlight the task of detecting small objects in aerial views, in particular for real-life low-resource scenarios~\cite{vedai, stanford,drones, dota,stanford,airbus}. The data distribution and computational constraints for such tasks often vastly differ from state-of-the-art benchmarks, for instance MS COCO~\cite{coco}.}\label{fig:intro} 
\end{figure}

In this work, we propose \textbf{\method{}}
 (\textbf{O}bject \textbf{D}etection with \textbf{G}rouped \textbf{I}nstances), 
a \emph{top-down}  detection scheme  specifically designed 
for efficiently handling inhomogeneous object distributions, while 
preserving detection performance.
%
Its key benefits and components are summarized as follows: 
\begin{itemize}  \setlength\itemsep{0em}
\item[(i)] a \emph{multi-stage pipeline}, in which each stage selects 
only \emph{a few promising regions} to be 
analyzed by the next stage, while discarding irrelevant image regions.
\item[(ii)] Fast single-shot detectors augmented with the ability to identify 
\emph{groups of objects} rather than just individual objects, 
thereby substantially reducing the number of regions that 
have to be considered.
\item[(iii)] \method{} reaches similar accuracies than ordinary single-shot detectors while operating at \textit{lower resolution} because groups of objects are generally larger and easier to detect than individual objects. This allows for a further reduction of computational requirements. 
\end{itemize}

%
We  present the proposed method, \method{}, and its training procedure in \hyperref[sec:methods]{Section \ref{sec:methods}}.
We then report main quantitative results as well as several ablation experiments in \hyperref[sec:exp]{Section \ref{sec:exp}}.
 	

\section{Related work}
\label{sec:related}
%
%

\medskip\noindent\textbf{Cascaded object detection.}
A popular  approach to  object detection consists in extracting
 numerous region  {\it proposals} and then classifying them as one of the object categories of
interest.
This includes models such as RFCN~\cite{RFCN},
RCNN and variants~\cite{fast_rcnn, rcnn, cascadercnn}, 
or SPPNet~\cite{SPPNet}. 
Proposal-based methods are very effective and can handle inhomogeneously
distributed objects, but are usually too slow for 
real-time usage, due to the large amount of proposals generated. 
Furthermore, with the exception of~\cite{faster_rcnn},
the proposals are generally class-independent,
which makes these methods more suitable for general scene understanding
tasks, where one is interested in  a wide variety of
classes. When targetting a specific object
category, class-independent proposals are wasteful, as 
most proposal regions are irrelevant to the task.

\medskip\noindent\textbf{Single-shot object detection and Multi-scale pyramids.}
In contrast, single-shot detectors, 
such as SSD~\cite{ssd}, or YOLO~\cite{yolo, yolo2, yolo3},  
 split the image into a regular grid of regions 
and predict object bounding boxes in each 
grid cell.
These single-shot detectors are efficient and can be made fast enough
for real-time operation, but  only provide a good speed-versus-accuracy trade-off 
when the objects of interest are distributed homogeneously
on the grid. 
In fact, the grid size has to be chosen with worst case scenarios
in mind: in order to identify all objects, the grid resolution has to be
fine enough to capture all objects even in image regions with high object 
density, which might rarely occur, leading to numerous empty cells.
Furthermore, the number of operations scales quadratically with 
the grid size, hence precise detection of individual small objects 
in dense clusters is often mutually exclusive with fast operation.
Recent work \cite{retinanet, sniper, autofocus, ssd, avdnet, clusdet} proposes to additionally exploit multi-scale
feature pyramids to better detect objects across varying scales.  This helps mitigate the aforementioned problem but does not
suppress it, and, in fact, these models are still better tailored for dense object detection.

Orthogonal to this, \method{} focuses on making the best of the given input resolution
and resources and instead resort to grouping objects when individual small instances are
too hard to detect, following the paradigms that ``coarse predictions are better than none''.
These groups are then refined in subsequent stages if necessary for the task at hand.
%

\medskip\noindent\textbf{Speed versus accuracy trade-off.}
Both designs involve intrinsic speed-versus-accuracy 
trade-offs, see for instance~\cite{DBLP:journals/corr/HuangRSZKFFWSG016} for
a deeper discussion, that make neither of them entirely  satisfactory 
 for real-world challenges, such as controlling an 
autonomous drone~\cite{drones}, localizing all objects of 
a certain type in aerial imagery~\cite{airbus} or efficiently 
detecting spatial arrangements of many small objects~\cite{deepscores}.

Our proposed method, \method{}, falls into neither of these 
two designs, but rather combines the strength of both
 in a flexible multi-stage pipeline: 
 It identifies a small number of specific regions of interest,
 which can also be interpreted as 
a form of proposals, thereby concentrating most of its computations 
on important regions. 
Despite the sequential nature of the pipeline, each individual prediction stage
 is based on a coarse, low resolution, grid, and thus very efficient. 
\method{}'s design resembles classical detection cascades~\cite{li2015,RowleyBK98,viola},
but differs from them in that it does not sequentially refine 
classification decisions for individual boxes but rather refines 
the actual region coordinates.
As such, it is conceptually similar to techniques based on 
branch-and-bound~\cite{lampert2010efficient,lampert2009efficient}, 
or on region selection by reinforcement learning~\cite{zoomin}.
Nonetheless, it strongly differs from these on a technical level 
as it only requires minor modifications of existing object 
detectors and can be trained with standard backpropagation 
instead of discrete optimization or reinforcement learning.
Additionally, \method{} generates meaningful groups of objects as
intermediate representations,
which can potentially be useful for other visual tasks. 
For example, it was argued  in \cite{VisualPhrases}  that recurring 
group structures can facilitate the detection of individual 
objects in complex scenes.
Currently, however, we  only make use of the fact that groups 
are visually more salient and easier to detect than 
individuals, especially at low image resolution.

\section{\method{}: Detection with Grouped Instances}
\label{sec:methods}
\begin{figure*}[tb]
\centering \includegraphics[width=0.905\textwidth]{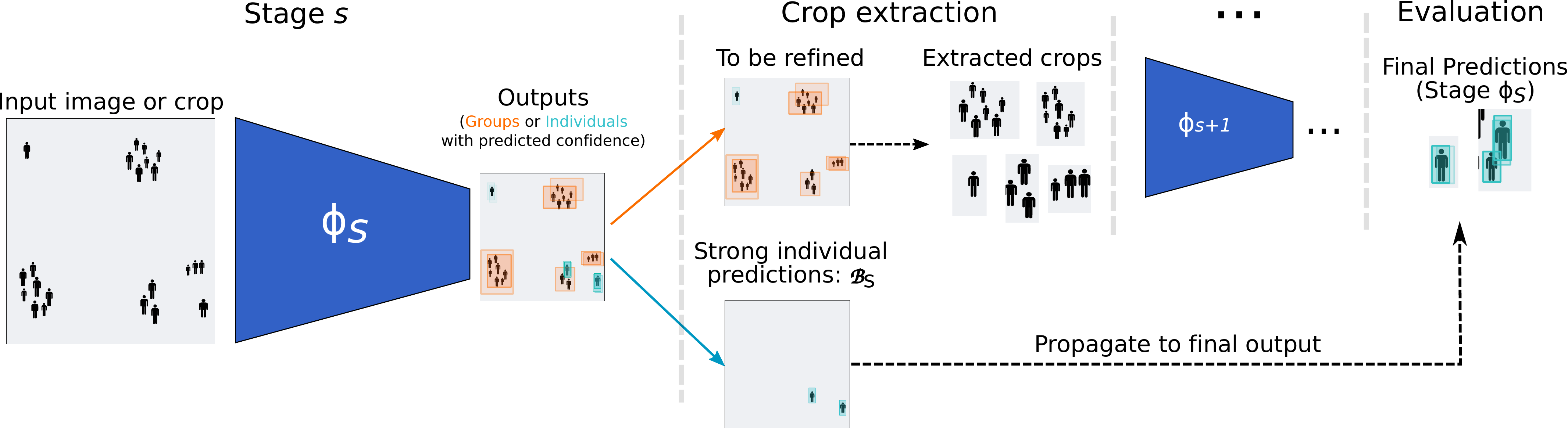}
\caption{\label{fig:main}Overview of  \method{}: Each stage $S$ consists of a single-shot detector that detects groups and individual objects, which are further processed to produce a few relevant image regions to be fed to subsequent stages and refine detections as needed. }
\end{figure*}

In \hyperref[sec:multistage]{Section \ref{sec:multistage}} 
we introduce the proposed multi-stage architecture and  the
notion of group of objects.
We then detail the training and
evaluation procedures in
\hyperref[subsec:training]{Section~\ref{subsec:training}}.

\subsection{Proposed architecture}\label{sec:multistage}

We design \method{} as a multi-stage detection architecture  
$\phi_S \circ \dots \circ \phi_1$, $S > 1$.
Each stage $\phi_s$ is a detection network, whose outputs can either be \emph{individual objects} 
or \emph{groups of objects}. 
In the latter case, the predicted bounding box defines 
a relevant image subregion, for which detections 
can be refined by feeding it as input  to the next 
stage.
To compare the model 
with standard detection systems, 
we  also constrain the last 
stage to only output individual objects.

\medskip\noindent\textbf{Grouped instances for detection.} We design each stage  as a lightweight neural network
that performs fast object detection. In our experiments, we build on standard single-shot detectors
such as  YOLO~\cite{yolo} or SSD~\cite{ssd}.
More precisely, $\phi_s$ consists of a \emph{fully-convolutional} network with output map $[I, J]$ directly proportional to the input image resolution. 
For each of the $I \times J$ cells in this uniform grid, the model predicts bounding boxes 
characterized by four coordinates --  the box center $(x,y)$, its width $w$ and height $h$,
and a predicted confidence score $c \in [0, 1]$.
Following common practice~\cite{yolo, yolo2, ssd}, we express the width and height
as a fraction of the total image width and height, while the coordinates of
the  center are parameterized  relatively to the cell it is linked to.
%
%
The confidence score $c$ is used for ranking the bounding boxes at inference time. 

For intermediate stages $s \leq S - 1$, we further incorporate 
the two following characteristics: \textit{First}, we augment
each predicted box with a binary \emph{group flag, $g$}, as
well as two real-valued \emph{offset values $(o_w,o_h)$}: 
The flag indicates whether the detector considers the 
prediction to be a single object, $g = 0$, or 
a group of objects, $g = 1$. 
The offset values  are used to appropriately rescale the stage outputs which are then
passed on to subsequent stages.
\textit{Second}, we design the intermediate stages to predict \textit{one bounding box} per cell.
This choice provides us with an \emph{intuitive definition of groups, which automatically adapts itself
to the input image resolution} without introducing additional hyperparamaters: 
If the model resolution $[I, J]$ is fine enough, there is at most one individual
object per cell, in which case the problem reduces to standard 
object detection. 
Otherwise, if a cell is densely occupied, then the model  resorts to
predicting one group enclosing the relevant objects. We provide further details on the group training process  in \hyperref[subsec:training]{Section~\ref{subsec:training}}.

\medskip\noindent\textbf{Multi-stage pipeline.}\label{sec:transition}
An overview of \method's multi-stage prediction 
pipeline is given in \hyperref[fig:main]{Figure \ref{fig:main}}. 

Each intermediate stage takes as inputs the outputs of the previous stage,
 which are processed to produce image regions in the following way: 
Let $B$ be a bounding box predicted at  stage $\phi_s$,
with confidence $c$ and binary group  flag $g$. 
We distinguish three possibilities: (i) the box can be 
discarded, (ii) it can be accepted as an individual object prediction, 
or (iii) it can be passed on to the next stage for further 
refinement. 
This decision is made based on two confidence thresholds, 
$\tau_{\text{low}}$ and $\tau_{\text{high}}$,
 leading to one of the three following actions:
\begin{itemize}\setlength\itemsep{0em}
\item[(i)] if $c \leq \tau_{\text{low}}$: The box $B$ is discarded.
\item[(ii)] if $c > \tau_{\text{high}}$ and $g = 0$: 
The box $B$ is considered a strong individual object candidate: 
we make it ``exit'' the pipeline and directly propagate it to the last stage's output as it is.
We denote the set of such boxes as ${\mathcal B}_s$.
\item[(iii)] if ($c >\tau_{\text{low}}$ and $g = 1$) 
or ($\tau_{\text{high}} \geq c > \tau_{\text{low}}$ and $g = 0$): 
The box $B$ is either a group or an individual with medium confidence and is 
 a candidate for refinement. 
\end{itemize}

After this filtering step, we apply non-maximum suppression (NMS) with threshold
$\tau_{\text{nms}}$ to the set of refinement candidates, in order to obtain (at most) 
 $\gamma_s$ boxes with high confidence and little overlap.
The resulting $\gamma_s$  bounding boxes are then processed to build the
image regions that will be passed on to the next stage 
by multiplying each box's width and height by $1/o_w$ and $1/o_h$, respectively, 
where $o_w$ and $o_h$ are the offset values learned by the detector. 

This rescaling step ensures  that the extracted patches
cover the relevant region well enough, and compensates for the fact
that the detectors are trained to \textit{exactly } predict ground-truth coordinates, 
rather than fully enclose them, hence sometimes underestimate the extent of the
relevant region.  
%
%
The resulting rescaled rectangular regions are  extracted from the 
input image and passed on as inputs to the next stage.
The final output of \method{} is the combination of object 
boxes predicted in the last stage, $\phi_S$, as well as the kept-back outputs 
from previous stages: ${\mathcal B}_1 \dots {\mathcal B}_{S-1}$.

The above patch extraction procedure can be tuned via four hyperparameters: 
$\tau_{\text{low}},\ \tau_{\text{high}},\ \tau_{\text{nms}},\ \gamma_s $.
At training time, we allow as many boxes to pass as the
memory budget allows. For our experiments, this was $\gamma^{\text{train}}_s=10$.
We also  do not use any of the aforementioned filtering during training, nor thresholding
($\tau^{\text{train}}_{\text{low}} = 0$, $\tau^{\text{train}}_{\text{high}} = 1$)
nor NMS ($\tau^{\text{train}}_{\text{nms}} = 1$) , because both negative and positive patches 
can be useful for training subsequent stages.
For test-time prediction we use a held-out validation set to determine their optimal values, 
as described in  \hyperref[cropextract]{Section \ref{cropextract}}.
Moreover, these hyperparameters can be easily changed on the fly, 
without retraining. This allows the model to easily adapt to 
changes of the input data characteristics, or  to make better use of 
an increased or reduced computational budget for instance.

\medskip\noindent\textbf{Number of stages.}
%
%
%
Appending an additional  refinement stage benefits the speed-vs-accuracy 
trade-off fit when the following  two criteria are met: 
\textit{First}, a low number of non empty cells; This 
 correlates to the number of extracted crops, thus to the number of 
feed-forward passes of subsequent stages. 
\textit{Second}, a small average group size: Smaller extracted regions lead to
 increased resolution once rescaled to the input size of the next stage, 
making the detection task which is fed to subsequent stages effectively easier. 

From the statistics reported in  
\hyperref[fig:intro]{Table \ref{fig:intro}}, we observe that 
 for classical benchmarks such as MS-COCO,  using only one stage suffices  
as groups are often  dense and cover large portions of the image:  
In that case, ODGI collapses to using a single-shot detector, such as~\cite{yolo, ssd}.
%
In contrast,  datasets of aerial views such as VEDAI~\cite{vedai} or SDD~\cite{stanford}
 contain small-sized group structures in large sparse areas. This is a typical scenario where 
the proposed refinement stages on groups improve the speed-accuracy trade-off.
We find that for the datasets used in our experiments 
$S = 2$ is sufficient, as regions extracted by the first stage 
typically exhibit a dense distribution of large objects.  
We expect the case $S>2$ to be beneficial for very large, 
\eg gigapixel images, but leave its study for 
future work.
Nonetheless, extending the model to this case should be straightforward: 
 This would introduce additional
hyperparameters as we have to tune the number of boxes $\gamma_s$ for 
each stage;  However, as we will see in the next section, these parameters
have little impact on training and can be easily tuned at test time.

\subsection{Training the model}\label{subsec:training}
We train each \method{} stage independently, using a combination of three
loss terms that we optimize with standard 
backpropagation (note that in the last stage of the pipeline, only the second term is active,
 as no groups are predicted):
\begin{align}
{\mathcal L}_{\method} = {\mathcal L}_{\text{groups}} + {\mathcal L}_{\text{coords}} + {\mathcal L}_{\text{offsets}}  \label{odgiloss}
\end{align}

${\mathcal L}_{\text{coords}}$ is a standard mean squares regression loss on the predicted coordinates
and confidence scores, as described for instance in \cite{yolo, ssd}.
The additional two terms are part of our contribution: 
The \emph{group loss}, ${\mathcal L}_{\text{groups}}$, 
 drives the model to classify outputs as individuals 
or groups, and the \emph{offsets loss}, ${\mathcal L}_{\text{offsets}}$, 
encourages better coverage of the extracted regions.
The rest of this section is dedicated to formally defining each loss term as well as
explaining how we obtain ground-truth coordinates for group bounding boxes.

\medskip\noindent\textbf{Group loss.}\label{grouploss}
Let $\mathbf{b} = b_{n=1 \dots N}$ be the original ground-truth individual bounding boxes.
We define $A^{ij}(n)$ as an indicator which takes value 1
\textit{iff} ground-truth box $b_n$ is assigned to output cell $(i,j)$ and 0 otherwise: 
\begin{align}
  A^{ij}(n) = \llbracket \left| b_n \cap \mbox{\small cell}_{ij} \right| > 0 \rrbracket ,  \mbox{\small with } \llbracket x \rrbracket = 1 \mbox{ \small if } x, \mbox{ \small else } 0
 \end{align}

For  the model  to predict groups of objects, we should in principle 
 consider  all the unions of subsets of $\mathbf{b}$ as potential targets.
 However, we defined our intermediate detectors to predict only one bounding box per cell by design, which allows us to avoid this combinatorial problem. Formally, let $B^{ij}$ be the predictor
 associated to cell $(i, j)$. We define its target ground-truth coordinates $\bar{B}^{ij}$
 and group flag $\bar{g}^{ij} $ as:
 \begin{align}\label{gtgroup}
   \bar{B}^{ij} &= \bigcup_{n | A^{ij}(n) = 1} b_n \\
   \bar{g}^{ij} &= \llbracket \# \{ n | A^{ij}(n) = 1 \} > 1 \rrbracket, 
   \end{align}
 with $\cup$ denoting the minimum enclosing bounding box of a set.
  We define  ${\mathcal L}_{\text{groups}}$ as a binary classification objective:

\begin{align}\label{group} 
  {\mathcal L}_{\text{groups}} = -  \sum_{i, j} &A^{i j}  \Big( \bar{g}^{j} \log(g^{ij} )
  \\
  &+  (1 - \bar{g}^{ij})  \log(1 - g^{ij}) \Big), \nonumber
\end{align}

where $A^{ij} = \llbracket \sum_n A^{ij}(n) > 0\rrbracket$ denotes whether cell $(i, j)$ is empty or not.
In summary, we build ground-truth $\bar{B}^{ij}$ and $\bar{g}^{ij}$ as follows:  
For each cell $(i,j)$, we build the set  $G^{ij}$ which ground-truth boxes $b_n$ of 
ground-truth boxes it intersects with.
%
%
If the set is non empty and only a single object box, $b$, falls into 
this cell, we set $\bar{B}^{ij} = b$ and $\bar{g}^{ij}= 0$. 
Otherwise,  $|G^{ij}|>1$ and we define $\bar{B}^{ij}$ as the union of bounding boxes 
in $G^{ij}$ and 
set $\bar{g}^{ij} = 1 $.
In particular, this  procedure  automatically adapts to the  resolution  $[I,J]$ in
a data-driven way, and can be implemented as a pre-processing step, thus 
does not produce any overhead at training time. 

\medskip\noindent\textbf{Coordinates loss.}\label{coordsloss}
Following the definition of target bounding boxes $\bar{B}^{ij}$ in \hyperref[gtgroup]{(\ref{gtgroup})},
we define the coordinates loss as a standard regression objective on the box coordinates and confidences, similarly to existing detectors~\cite{rcnn, fast_rcnn, ssd, multibox, yolo}.

\begin{align}
\label{obj}
  {\mathcal L}_{\mbox{\tiny coords}} = \!\!\sum_{i, j} A^{ij} \big( &\| B^{ij} - \bar{B}^{ij} \|^2 + \omega_{\mbox{\tiny conf}}\ \| c^{ij} - \bar{c}^{ij}\|^2 \nonumber \\ &+ \omega_{\mbox{\tiny no-obj}}  \sum_{i, j} (1 - A^{ij})  \left( c^{ij} \right)^2 \big)\\
  \bar{c}^{ij} = \iou{} (B^{ij}&, \bar{B}^{ij}) = \frac{|B^{ij} \cap \bar{B}^{ij} |}{|B^{ij} \cup \bar{B}^{ij} |}
\end{align}
The first two terms are ordinary  least squares regression objectives between the predicted coordinates and confidence scores and their respective assigned ground-truth. The ground-truth for the confidence score is defined as the intersection over union (\iou{}) between the corresponding prediction and its assigned target.
Finally, the last term in the sum is a weighted \emph{penalty term} to push confidence scores for empty cells towards zero. In practice, we use the same weights as in~\cite{yolo}, i.e. $\omega_{\text{conf}}=5$ and $\omega_{\text{no-obj}}=1$.

\medskip\noindent\textbf{Offsets loss.} \label{sec:offsets} 
In intermediate stages, \method{} predicts offset values for each box, $o_w$ 
and $o_h$,  that are used to rescale the region of interest 
when it is passed as input to the next stage,
as described in \hyperref[sec:transition]{Section \ref{sec:transition}}. 
The corresponding predictors are trained using the following \emph{offsets loss}: 
\begin{align}
\label{offsets}
{\mathcal L}_{\text{offsets}} = \sum_{i, j}  &A^{ij} \Big[ \left( o_w - \bar{o}_w(B^{ij}, \bar{B}^{ij})  \right)^2 \nonumber\\
+ &\left( o_h - \bar{o}_h(B^{ij}, \bar{B}^{ij})  \right)^2  \Big]. 
\end{align}
The target values, $\bar{o}_h(B^{ij}, \bar{B}^{ij})$ and $\bar{o}_w(B^{ij}, \bar{B}^{ij})$,
for vertical and horizontal offsets, are determined as follows:
First, let $\alpha$ denote the center y-coordinate and $h$ the height. 
Ideally, the vertical offset should cause the rescaled version of 
$B^{i j}$ to encompass both the original $B^{i j}$ and its assigned ground-truth 
box $\bar{B}^{ij}$ with a certain margin $\delta$, which we set to half the average 
object size ($\delta=0.0025$). 
Formally: 
\begin{align}
\label{eq:offset}
h^{\text{scaled}}(B, \bar{B}) &= \max ( |(\alpha(\bar{B})\!+\!h(\bar{B}) / 2 + \delta)\!-\!\alpha(B)|, \nonumber \\
 &\ \ \ \ \ \ \ \ \ \ \ \ \ \ |(\alpha(\bar{B})\! -\! h(\bar{B}) / 2 - \delta)\! -\! \alpha(B)| )\nonumber\\
\bar{o}_h(B^{ij}, \bar{B}^{ij}) &= \max(1, h(B^{ij}) / h^{\text{scaled}}(B^{ij}, \bar{B}^{ij}))
\end{align}
For the horizontal offset, we do the analogous construction using the $B^{ij}$'s center x-coordinate 
and its width instead.


\medskip\noindent\textbf{Evaluation metrics.}
\label{subsec:eval}
We quantitatively evaluate the ODGI pipeline as a 
standard object detector: 
Following the common protocol from PASCAL VOC 2010 and 
later challenges~\cite{pascal}, we sort the list of predicted  
boxes in decreasing order of confidence score and compute the 
\emph{average precision (\map{})} respectively to the
ground-truth, at the \iou{} cut-offs 
of 0.5 (\emph{standard}) and 0.75 (\emph{more precise}). 
In line with our target scenario of single-class object 
detection, we ignore class information in experiments and focus 
on raw detection.
Class labels could easily be added, either on the level of individual
box detections, or as a post-processing classification operation,
which we leave for future work.

\medskip\noindent\textbf{Multi-stage training.}
\label{subsec:end-to-end}
By design, the inputs of stage $s$ are obtained from the outputs of stage $s - 1$. However it is cumbersome to wait for each stage to be fully trained before starting to train the next one. In practice we notice that even after only a few epochs, the top-scoring predictions of intermediate detectors often detect image regions that can be useful for the subsequent stages, thus we propose the following training procedure: After $n_e = 3$ epochs of training the first stage, we start training the second, querying new inputs from a queue  fed by the outputs of the first stage.
This allows us to jointly and efficiently train the two stages, and this delayed training scheme works well in practice.


\section{Experiments}
\label{sec:exp}

\begin{figure*}[tb]
  \begin{center}
    \includegraphics[width=0.33\textwidth]{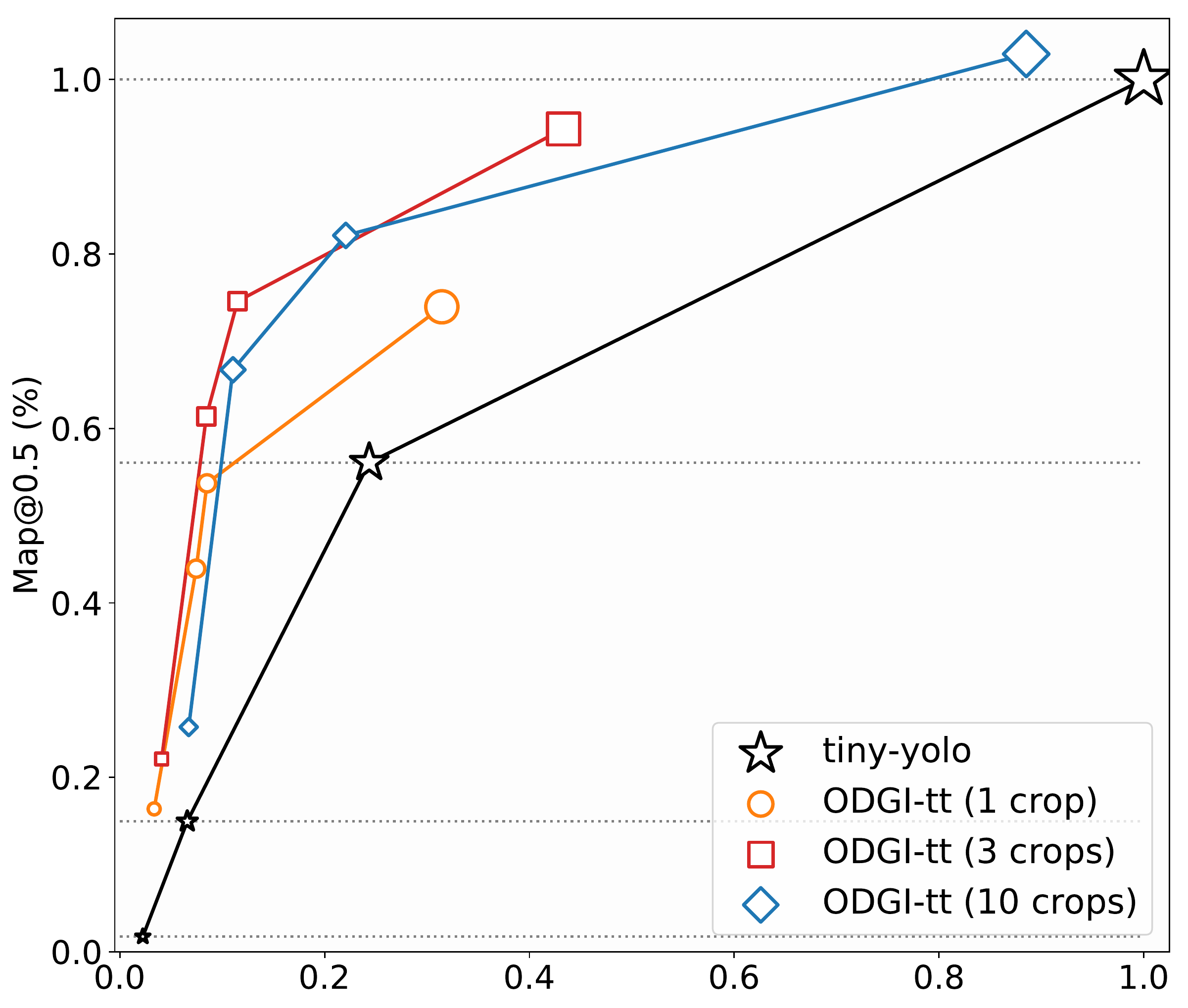}
    \includegraphics[width=0.33\textwidth]{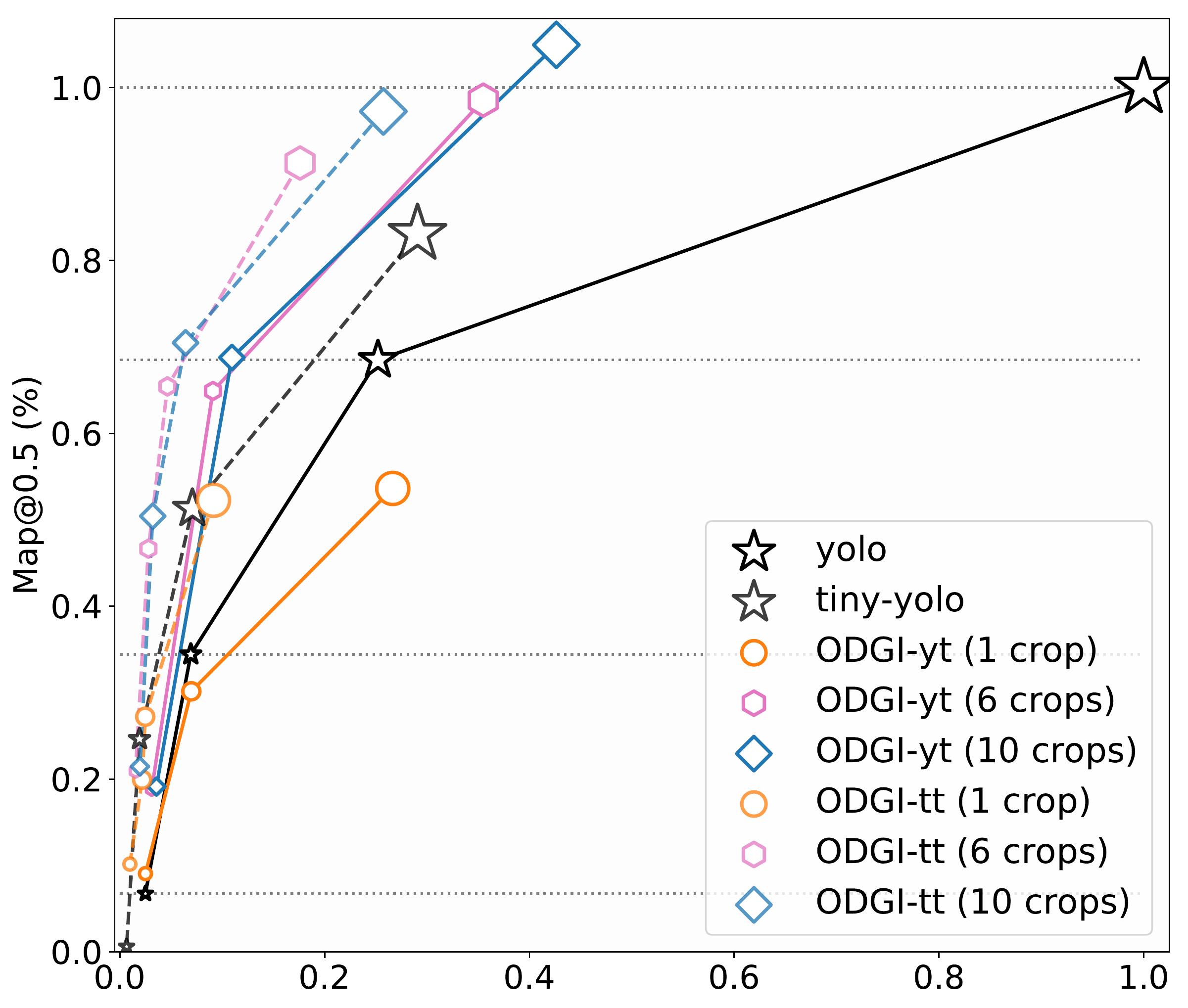}
    \includegraphics[width=0.33\textwidth]{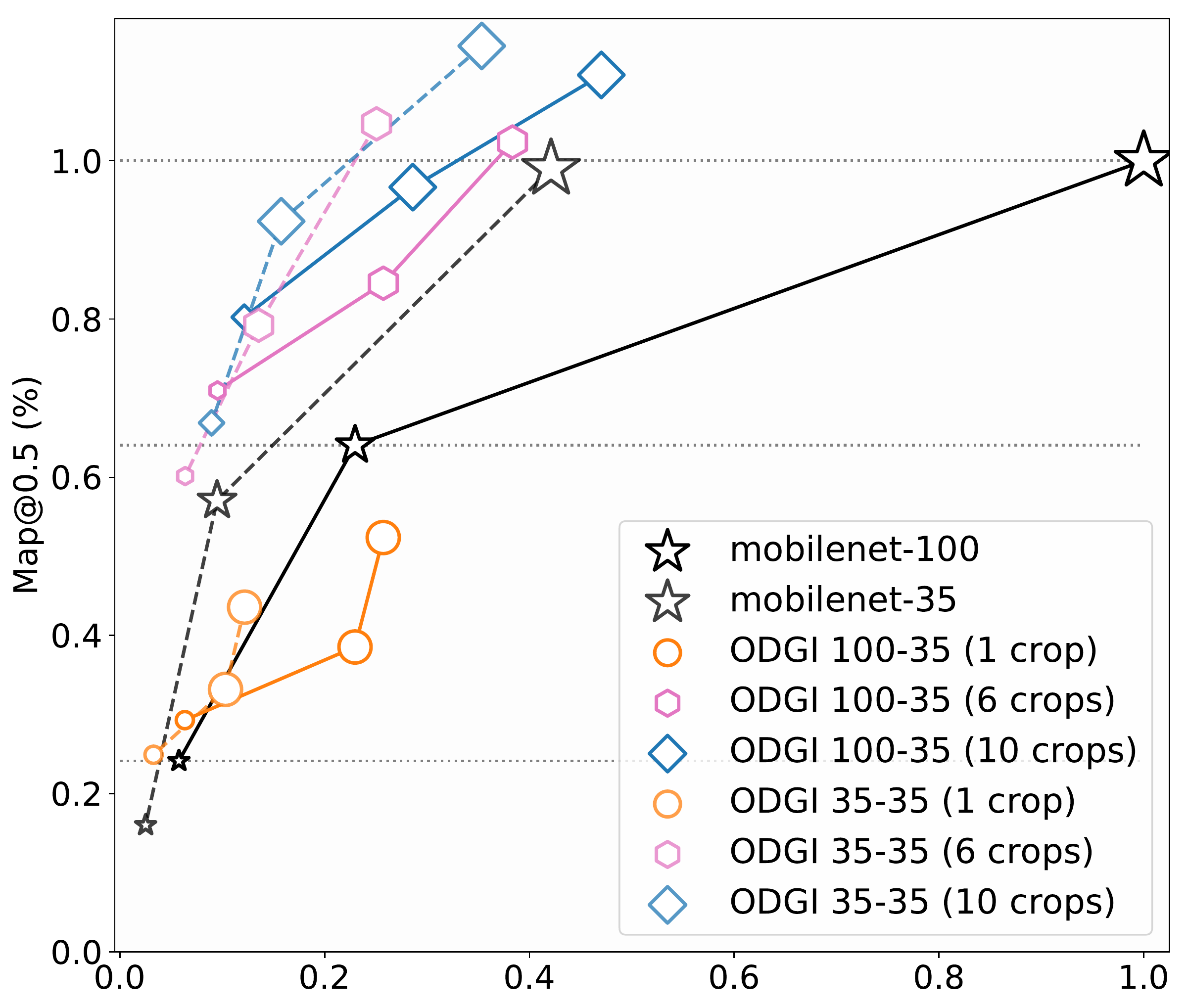}
  \end{center}
  \vspace{-0.4cm}
  \caption{\label{fig:mainplot}Plots of \map{@0.5} versus runtime (CPU) for the \vedai{} and \sdd{} datasets on three different backbone architectures. The metrics are reported as percentages relative to the baseline run at full resolution.
    Each marker corresponds to a different input resolution, which the marker size is proportional to.
    The black line represents the baseline model, while each colored line corresponds to a specific number of extracted crops, $\gamma_1$, For readability, we only report results for a subset of $\gamma_1$ values, and provide full plots in the supplemental material. 
  }
\end{figure*}

\label{models}
We report  experiments on two aerial views datasets: 
VEDAI~\cite{vedai} contains 1268 aerial views of countryside and 
city roads for vehicle detection. 
Images are 1024x1024 pixels and contain on average 2.96 objects of interest.
We perform 10--fold cross validation, as  in~\cite{vedai}.  For each run, we use 8 folds for training, one for validation and one for testing. All reported metrics are averaged over the 10 runs.
Our second benchmark, SDD~\cite{stanford}, 
contains drone videos taken at different locations with bounding box annotations of road users.
To reduce redundancy, we extract still images every 40 frames, which we then pad and resize to 1024x1024 pixels to compensate for different aspect ratios.
For each location, we perform a random train/val/test split with 
ratios 70\%/5\%/25\%, resulting in total in 9163, 651 and 3281 images respectively. 
On average, the training set contains 12.07 annotated objects per image.
\sdd{}  is overall much more challenging than \vedai{}:  at full 
resolution, objects are small and hard to detect, even to the human eye. 

%
We consider three common backbone networks for \method{} and baselines:
\texttt{tiny}, a simple 7-layer fully convolutional network
based on the tiny-YOLO architecture, \texttt{yolo}, a VGG-like
network similar to the one used in YOLOv2~\cite{yolo} and finally \texttt{MobileNet}~\cite{mobilenet}, which is for instance
used in SSD Lite~\cite{ssd}.
More specifically, on the \vedai{} dataset, we train a standard \texttt{tiny-yolov2} detector
as  baseline and compare it to \method-\emph{teeny-tiny} (\ott), which refers to two-stage
ODGI with \texttt{tiny} backbones.
For \sdd, objects are much harder to detect, thus we use a stronger \emph{\yolo{} V2} model as baseline.
We compare this to \method-\emph{teeny-tiny} as above as well a stronger variant, \method-\emph{yolo-tiny} (\oyt),
in which $\phi_1$ is based on the \texttt{yolo} backbones and $\phi_2$ on \texttt{tiny}.
Finally we also experiment with the lightweight \texttt{MobileNet} architecture as baseline and backbones, with depth multipliers 1 and 0.35. The corresponding ODGI models are denoted as \method-\textit{100}-\textit{35} and  \method-\textit{35}-\textit{35}.
All models are trained and evaluated at various resolutions to investigate different grouping scenarios.
In all cases, the detector grid size scales linearly with the 
image resolution, because of the fully convolutional network structures, 
ranging from a $32\times 32$ grid for 1024px inputs to $2\times 2$ for 64px. 

We implement all models in Tensorflow and train with the Adam optimizer~\cite{adam}
and learning rate 1{\small e-3}.
To facilitate reproducibility, we make our code publicly available~\footnote{Github repository, \url{https://github.com/ameroyer/ODGI}}.

\subsection{Main results}
\label{sec:mainresults}

To benchmark detection accuracy, we evaluate the average precision (\map{}) 
for the proposed \method{} and baselines.
As is often done, we also apply     
non-maximum suppression  to the final predictions, with \iou{} threshold of 0.5 and no limit on the number of outputs, to remove near duplicates for all methods. 
Besides retrieval performance, we assess the computational and memory resource requirements of the different methods:   
%
We record the number of boxes predicted by each model, 
and measure the average runtime of our implementation for one forward pass 
  on a single image. 
As reference hardware, we use a server with 
\emph{2.2 GHz Intel Xeon processor (short: CPU)}
in single-threaded mode. 
Additional timing experiments on weaker and stronger hardware, 
as well as a description of 
how we pick \method{}'s test-time hyperparameters 
can be found in \hyperref[sec:extra]{Section \ref{sec:extra}}.

\begin{figure*}[t]\small\centering
  \def \exampleid {5}
  \def \trimleft {11.8cm}
  \def \trimbot {8.3cm}
  \def \trimright {0cm}
  \def \trimtop {5.5cm}
  \def \imgwidth {.275\textwidth}
  \begin{tabular}{ccc}
    \includegraphics[width=\imgwidth, trim={\trimleft{} \trimbot{} \trimright{} \trimtop{}}, clip]{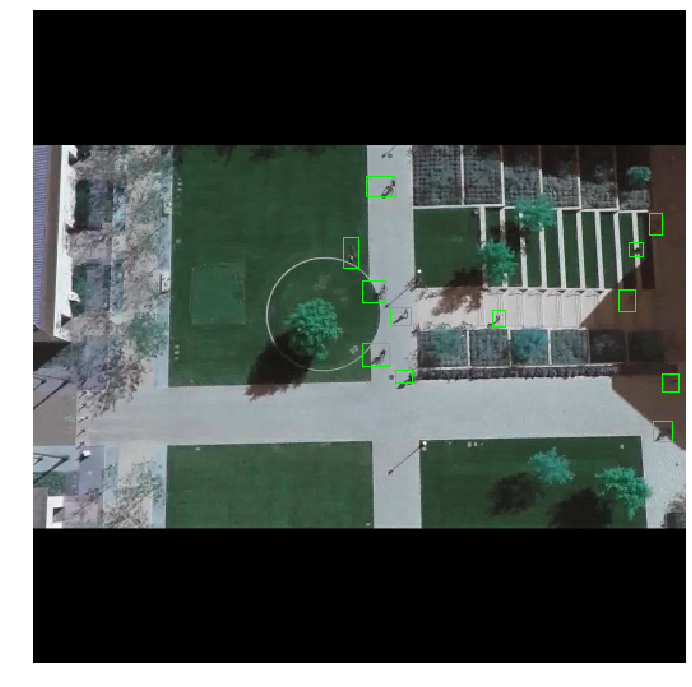}
    &
      \includegraphics[width=\imgwidth, trim={\trimleft{} \trimbot{} \trimright{} \trimtop{}}, clip]{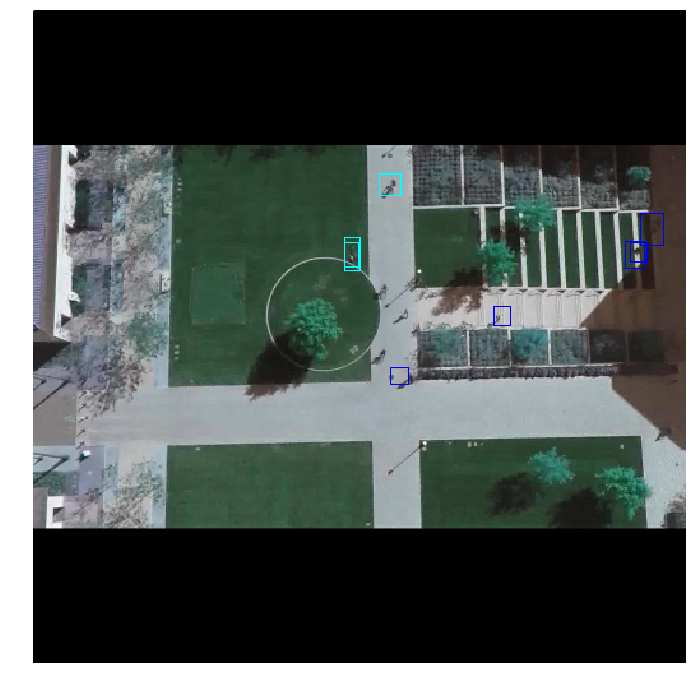}
    &
      \includegraphics[width=\imgwidth, trim={\trimleft{} \trimbot{} \trimright{} \trimtop{}}, clip]{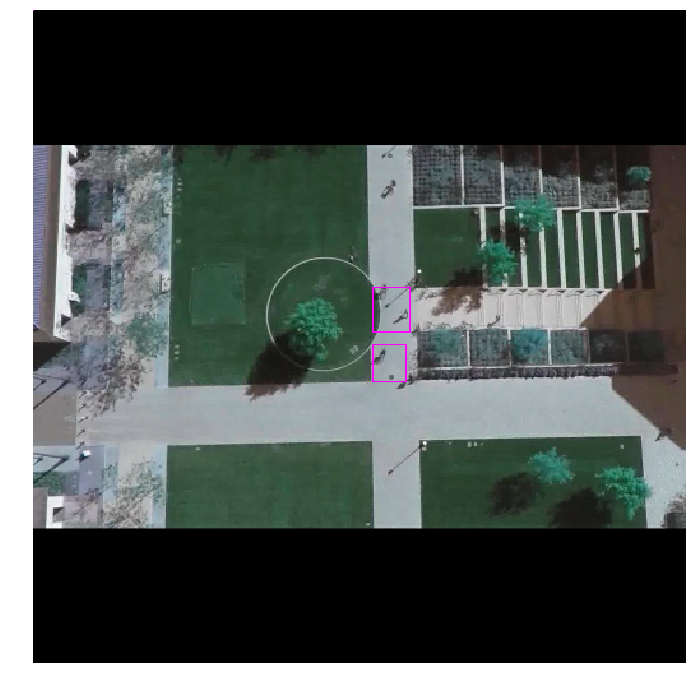}
    \\ 
    (a) Ground-truth & \makecell{(b) \textbf{stage 1}: individual boxes (\textcolor{cyan}{cyan}: $c \geq \tau_{\mbox{\tiny high}}$)} & (c) \textbf{stage 1}: detected group boxes
    \\
    \includegraphics[width=\imgwidth, trim={\trimleft{} \trimbot{} \trimright{} \trimtop{}}, clip]{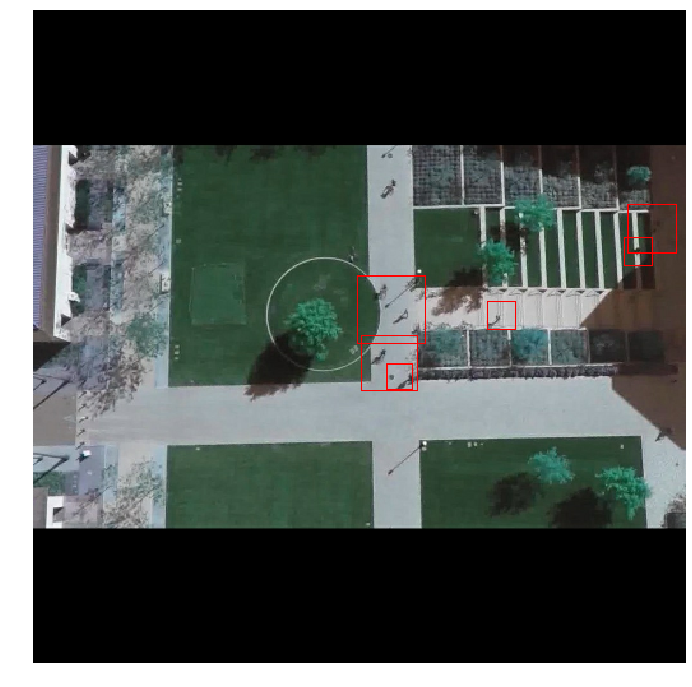}
    &
      \includegraphics[width=\imgwidth, trim={\trimleft{} \trimbot{} \trimright{} \trimtop{}}, clip]{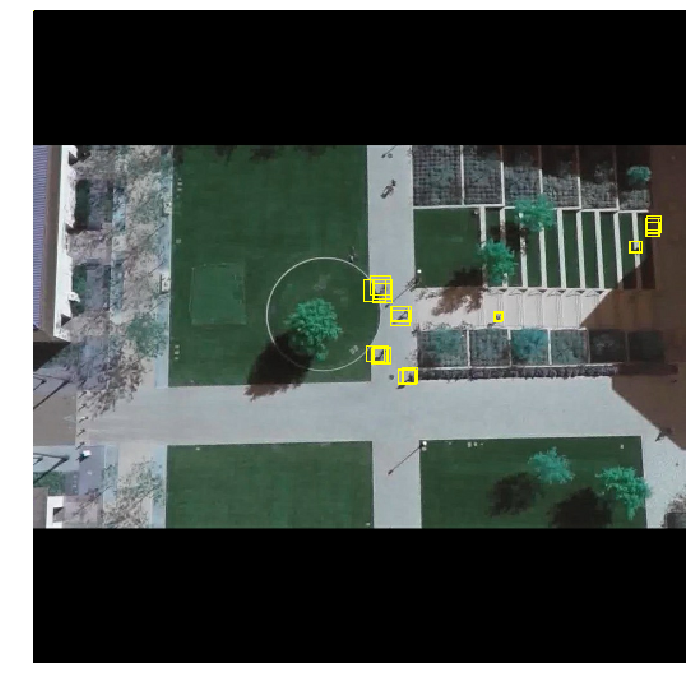}
    &
      \includegraphics[width=\imgwidth, trim={\trimleft{} \trimbot{} \trimright{} \trimtop{}}, clip]{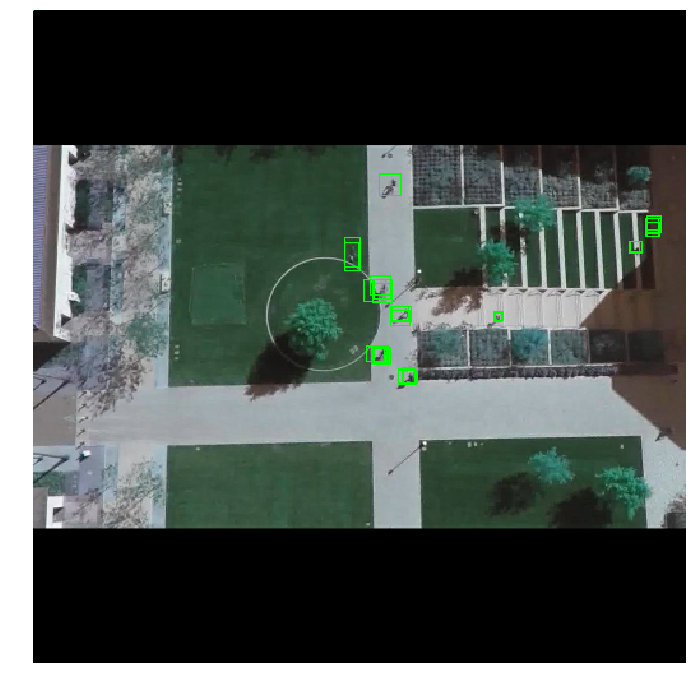}
    \\
    (d) \textbf{stage 1}: regions passed to stage 2 & (e) \textbf{stage 2}: detected object boxes & (f) \method{}: overall detected object boxes
  \end{tabular}
  \caption{\label{fig:qual}Qualitative results for \method{}. No filtering step was applied here, but for readability we only display boxes predicted with confidence at least 0.5. 
    Best seen on PDF with zoom. Additional figures are provided in the supplemental material.}
\end{figure*}

We report experiment results in \hyperref[fig:mainplot]{Figure \ref{fig:mainplot}} (see \hyperref[tab:main]{Table~\ref{tab:main}} for exact numbers).
We find that the proposed method improves over standard single-shot detectors in 
two ways: \textit{First}, when comparing models with similar accuracies, \method{} 
generally requires fewer 
evaluated boxes and shorter runtimes, and often lower input image resolution. 
In fact, only a few relevant regions are passed to the 
second stage, at a smaller input resolution, hence they incur a small 
computational cost, yet the ability to selectively refine the boxes can
substantially improve detection. 
\textit{Second}, for any given input resolution, \method{}'s refinement cascade generally 
improves detection retrieval, in particular at lower resolutions, \eg 256px: 
In fact, \method{}'s first stage can be kept efficient and operate 
at  low resolution, because the regions it extracts do not 
have to be very precise.
Nonetheless, the regions selected in the first stage form an easy-to-solve detection 
task for the second stage (see for instance
\hyperref[fig:qual]{Figure \ref{fig:qual} (d)}), which leads to more  precise
detections after refinement.
This also motivates our choice of mixing backbones, \eg using \method-\emph{yolo-tiny}, 
 as detection in stage 2 is usually much easier. 

\begin{table}[b]
  \resizebox{\linewidth}{!}{%
    \begin{tabular}{|c||c|c||c||c|}
      \hline
      \small \textbf{\vedai{}} & \small \map{\footnotesize @}\footnotesize0.5 & \map{\footnotesize @}\footnotesize0.75 &\small CPU [s] & \small \#\textsubscript{boxes}\\
      \hline
      \hline 
      \footnotesize \ott  512-256   &        0.646     &    0.422    &      0.83    &        $\leq$ 448 \\ 
      \footnotesize \ott  512-64    &         0.562    &   0.264     &      0.58    &       $\leq$ 268  \\ 
      \footnotesize \ott  256-128   &         0.470    &   0.197     &      0.22    &       $\leq$ 96    \\ 
      \footnotesize \ott 256-64     &         0.386    &   0.131     &     0.16     &        $\leq$ 72   \\  
      \footnotesize \ott   128-64   &        0.143     &     0.025   &      0.08    &       $\leq$ 24   \\ 
      \hline
      \footnotesize tiny-yolo 1024    &        0.684     &   0.252     &      1.9    &        1024           \\
      \footnotesize tiny-yolo  512    &         0.383    &    0.057    &     0.47    &        256             \\
      \footnotesize tiny-yolo  256    &         0.102    &   0.009     &      0.13   &       64                \\
      \hline
    \end{tabular}
  }

  \vspace{0.4cm}     
  \resizebox{\linewidth}{!}{%
    \begin{tabular}{|c||c|c||c||c|}
      \hline
      \small \textbf{\sdd{}} & \small \map{\footnotesize @}\footnotesize0.5 & \map{\footnotesize @}\footnotesize0.75 &\small CPU [s] & \small \#\textsubscript{boxes}\\
      \hline
      \hline
      \footnotesize \oyt 512-256     &0.463      &      0.069  &     2.4       & $\leq$ 640 \\ 
      \footnotesize \ott 512-256     &0.429      &      0.061  &     1.2       & $\leq$ 640\\ 
      \footnotesize \oyt 256-128     &0.305      &      0.035  &     0.60      & $\leq$  160 \\ 
      \footnotesize \ott 256-128     &0.307      &      0.044  &     0.31      & $\leq$ 160\\ 
      \hline
      \footnotesize yolo 1024          &0.470      &   0.087     &     6.6       & 1024 \\
      \footnotesize yolo 512           &0.309      &  0.041      &     1.7       & 256 \\
      \footnotesize yolo 256           &0.160      & 0.020       &    0.46       & 64 \\
      \hline
    \end{tabular}}

  \vspace{0.4cm}    
  \resizebox{\linewidth}{!}{%
    \begin{tabular}{|c||c|c||c||c|}
      \hline
      \small \textbf{\sdd{}} & \small \map{\footnotesize @}\footnotesize0.5 & \map{\footnotesize @}\footnotesize0.75 &\small CPU [s] & \small \#\textsubscript{boxes}\\
      \hline
      \hline
      \footnotesize \omb 512-256     & 0.434    & 0.061    & 0.76         & $\leq$ 640 \\ 
      \footnotesize \omb 256-128    & 0.294     & 0.036     & 0.19    & $\leq$ 160\\ 
      \hline
      \footnotesize \mobilenet-100 1024    & 0.415     & 0.061     & 1.9    &  1024  \\
      \footnotesize \mobilenet-100 512     & 0.266     & 0.028     & 0.46     & 256 \\
      \footnotesize \mobilenet-100 256     & 0.100     & 0.009     & 0.12     & 64 \\
      \hline
    \end{tabular}}

  \vspace{0.4cm}   
  \resizebox{\linewidth}{!}{%
    \begin{tabular}{|c||c|c||c||c|}
      \hline
      \small \textbf{\sdd{}} & \small \map{\footnotesize @}\footnotesize0.5 & \map{\footnotesize @}\footnotesize0.75 &\small CPU [s] & \small \#\textsubscript{boxes}\\
      \hline
      \hline
      \footnotesize \oms 512-256     &  0.425     & 0.055     & 0.50       & $\leq$ 640 \\ 
      \footnotesize \oms 256-128    &  0.250     & 0.029     & 0.13   & $\leq$ 160\\ 
      \hline
      \footnotesize \mobilenet-35 1024    &  0.411    & 0.054     & 0.84     &  1024  \\
      \footnotesize \mobilenet-35 512     & 0.237     & 0.026     & 0.19     & 256 \\
      \footnotesize \mobilenet-35 256     & 0.067     & 0.007     & 0.050     & 64 \\
      \hline
    \end{tabular}}
\vspace{0.05cm}
  \caption{\map{} and timing results on the VEDAI and SDD datasets for the model described in \hyperref[models]{Section \ref{models}}. The results for \method{} models are reported with $\gamma^{\text{test}}_1$ chosen as described in \hyperref[cropextract]{Section \ref{cropextract}}. 
}\label{tab:main}
\end{table}

\begin{table*}[t]
  \resizebox{.515\linewidth}{!}{%
    \begin{tabular}{|c||c|c|c|c||c|c|c|}
      \hline
      \multirow{ 2}{*}{\textbf{\vedai{}}} &  \footnotesize   \ott   &      \footnotesize \ott  &  \footnotesize \ott   &  \footnotesize \ott &    \footnotesize tiny-yolo  & \footnotesize tiny-yolo       &  \footnotesize tiny-yolo  \\
                        & \footnotesize 512-256 &      \footnotesize 256-128 &   \footnotesize 256-64 &  \footnotesize 128-64 & \footnotesize 1024 &   \footnotesize 512   &     \footnotesize 256\\
      \hline
      \small \map{\footnotesize @}\footnotesize0.5  & 0.65 &  0.47 & 0.39 & 0.14 & 0.68 & 0.38  & 0.10 \\
      \hline
      \hline
      \small Raspi [s]  & 4.9 & 1.2    & 0.87     & 0.44    & 10.5     & 2.6     & 0.70    \\
      \hline
      \small GPU [ms]   & 13.9 & 11.7    & 11.7    & 11.7    & 14.3     & 8.2     & 7.0      \\
      \hline
      \hline
      \#\textsubscript{parameters} & 22M& 22M& 22M& 22M& 11M& 11M& 11M\\
      \hline
      \#\textsubscript{pixels}  & 458k  & 98k& 73k& 25k & 1M & 262k& 65k\\
      \hline
    \end{tabular}
  }
  \resizebox{.485\linewidth}{!}{%
    \begin{tabular}{|c||c|c|c|c||c|c|c|}
      \hline
      \multirow{ 2}{*}{\textbf{\sdd{}}} &  \footnotesize   \oyt   &  \footnotesize \ott   &   \footnotesize \oyt  &   \footnotesize \ott    &   \footnotesize yolo  &  \footnotesize yolo & \footnotesize yolo  \\
                      &  \footnotesize 512-256 &   \footnotesize 512-256 &    \footnotesize 256-128 &   \footnotesize 256-128  & \footnotesize 1024 &   \footnotesize 512 &   \footnotesize 256\\
      \hline
      \small \map{\footnotesize @}\footnotesize0.5  & 0.46 & 0.43  & 0.31 & 0.31 & 0.47 & 0.32 & 0.16 \\
      \hline
      \hline
      \small Raspi [s]  & 16.4     & 7.0     & 4.6     & 1.8    & 46.9    & 12.1     & 3.4 \\
      \hline
      \small GPU [ms]   & 24.5    & 14.7    & 18.7    & 12.0    & 34.7    & 16.9    & 13.5 \\
      \hline
      \hline
      \#\textsubscript{parameters} & 62M& 22M& 62M& 22M& 51M& 51M& 51M \\
      \hline
      \#\textsubscript{pixels}  & 655k & 655k & 164k & 164k & 1M & 260k & 65k\\
      \hline
    \end{tabular}
  }

  \vspace{0.08cm}
  \resizebox{.51\linewidth}{!}{%
    \begin{tabular}{|c||c|c||c|c|c|}
      \hline
      \multirow{ 2}{*}{\textbf{\sdd{}}} &  \footnotesize   \omb   &  \footnotesize \omb   &  \footnotesize \mobilenet-100  &  \footnotesize \mobilenet-100 & \footnotesize \mobilenet-100  \\
                      &  \footnotesize 512-256 &    \footnotesize 256-128 &   \footnotesize 1024 &   \footnotesize 512 &   \footnotesize 256\\
      \hline
      \small \map{\footnotesize @}\footnotesize0.5  & 0.43     & 0.29     & 0.42     & 0.27     & 0.10\\
      \hline
      \small Raspi [s]     & 6.6     & 1.6    & 17.3     & 4.0     & 0.92\\
      \hline
      \small GPU [ms]    & 19.9    & 17.6    & 23.1    & 11.0     & 9.5 \\
      \hline
      \hline
      \#\textsubscript{parameters} & 2.6M & 2.6M & 2.2M & 2.2M & 2.2M \\
      \hline
      \#\textsubscript{pixels}  & 655k & 164k & 1M & 260k & 65k\\
      \hline
    \end{tabular}
    }
    ~
  \resizebox{.478\linewidth}{!}{%
    \begin{tabular}{|c||c|c||c|c|c|}
      \hline
      \multirow{ 2}{*}{\textbf{\sdd{}}} &  \footnotesize   \oms   &  \footnotesize \oms   &  \footnotesize \mobilenet-35  &  \footnotesize \mobilenet-35 & \footnotesize \mobilenet-35  \\
                      &  \footnotesize 512-256 &    \footnotesize 256-128 &   \footnotesize 1024 &   \footnotesize 512 &   \footnotesize 256\\
      \hline
      \small \map{\footnotesize @}\footnotesize0.5  & 0.43     & 0.25     & 0.41     & 0.24     & 0.067 \\
      \hline
      \small Raspi [s]     & 4.1     & 1.0     & 6.8     & 1.5     & 0.42 \\
      \hline
      \small GPU [ms]    & 17.8    & 17.4    & 13.9     & 9.8     & 9.3 \\
      \hline
      \hline
      \#\textsubscript{parameters} & 800k & 800k & 400k & 400k & 400k\\
      \hline
      \#\textsubscript{pixels}  & 655k & 164k & 1M & 260k & 65k\\
      \hline
    \end{tabular}
  }

  \caption{Additional timing results. 
    Time is indicated in seconds for a \emph{Raspberry Pi (Raspi)}, and in milliseconds 
    for an \emph{Nvidia GTX 1080Ti graphics card (GPU)}. \#\textsubscript{pixels} is
    the total number of pixels processed and \#\textsubscript{parameters}, the number of model parameters.}\label{tab:res}
\end{table*}

\subsection{Additional Experiments}

\medskip\noindent\textbf{Runtime.}\label{sec:extra}
Absolute runtime values always depend 
on several factors, in particular the software 
implementation and hardware. 
In our case, software-related differences are not an issue, as all models rely
on the same core backbone implementations.
%
%
To analyze the effect of hardware, we performed additional experiments on weaker 
hardware, a \emph{Raspberry Pi 3 Model B with 1.2 GHz ARMv7 CPU (Raspi)}, 
as well as stronger hardware, an \emph{Nvidia GTX 1080Ti graphics card (GPU)}.
\hyperref[tab:res]{Table \ref{tab:res}} shows the resulting 
runtimes of one feed-forward pass for  the same models and baselines as in  \hyperref[tab:main]{Table \ref{tab:main}}.
We also report the total number of pixels processed 
by each method, \ie that have to 
be stored in memory during one feed-forward pass, as well 
as the number of parameters.

The main observations of the previous section 
again hold: 
On the Raspberry Pi, timing ratios are roughly the same as on the Intel CPU, 
only the absolute scale changes.
The differences are smaller on GPU, but \method{} is 
still faster than the baselines in most cases at similar 
accuracy levels.
Note that for the application scenario we target, the GPU timings 
are the least representative, as systems operating under 
resource constraints typically cannot afford the usage of a 250W 
graphics card (for comparison, the Raspberry Pi has a power 
consumption of approximately  1.2W).

\medskip\noindent\textbf{Hyperparameters.}\label{cropextract}
%
%
As can be seen in \hyperref[fig:mainplot]{Figure \ref{fig:mainplot}}, a higher number of crops, $\gamma^{\text{test}}_1$, improves detection, but comes at a higher computational cost. Nonetheless, \method appears to have a better accuracy-speed ratio for most values of $\gamma^{\text{test}}_1$. 
For practical purposes, we suggest to choose $\gamma^{\text{test}}_1$ based on how many patches are effectively used for detection. 
We define the \textit{occupancy rate} of a crop as the sum of the intersection ratios of ground-truth boxes that appear in this crop.
We then say a crop is \textit{relevant} if it has a non-zero occupancy rate, \ie it contains objects of interest: 
For instance, at input resolution  512px on \vedai{}'s validation set, we obtain an average of 2.33 relevant crops , hence we set $\gamma^{\text{test}}_1 = 3$.
%
%
The same analysis on \sdd{} yields  $\gamma^{\text{test}}_1 = 6$. 
%

Three additional hyperparameters influence \method{}'s behavior: 
$\tau^{\text{test}}_{\text{low}}$, $\tau^{\text{test}}_{\text{high}}$, and $\tau^{\text{test}}_{\text{nms}}$,
all of which appear in the patch extraction pipeline.
For a range of $\gamma_1 \in [1, 10]$, and for each input resolution, 
we perform a parameter sweep on the held-out validation set over the ranges 
$\tau_{\text{low}} \in \{ 0., 0.1, 0.2, 0.3, 0.4\}$, $\tau_{\text{high}} \in \{ 0.6, 0.7, 0.8, 0.9, 1.0\}$, and $\tau_{\text{nms}} \in \{ 0.25, 0.5, 0.75\}$.
Note that network training is independent from these parameters as discussed 
in \hyperref[sec:transition]{Section \ref{sec:transition}}. Therefore the 
sweep can be done efficiently using pretrained $\phi_1$ and $\phi_2$,
changing only the patch extraction process.
We report full results of this validation process in the supplemental material. 
The main observations are as follows: 

(i) $\tau^{\text{test}}_{\text{low}} $ is usually in $\{0, 0.1\}$. 
 This indicates that the low confidence patches 
are generally true  negatives that need not be filtered out. 
%
(ii) $\tau_{\text{high}} \in \{ 0.8, 0.9\}$ for \vedai{} and  $\tau_{\text{high}} \in \{ 0.6, 0.7\}$ for \sdd{}.
This reflects intrinsic properties of each dataset: \vedai{} images contain only few objects 
which are easily covered by the extracted crops. It is always beneficial to 
refine these predictions, even when they are individuals with high confidence, hence a high value of $\tau_{\text{high}}$. 
In contrast, on the more challenging \sdd{}, \method{} more often uses the shortcut 
for confident individuals in stage 1, in order to focus the refinement stage on groups and 
lower-confidence individuals which can benefit more.
(iii) $\tau^{\text{test}}_{\text{nms}}$ is usually equal to 0.25, which encourages 
non-overlapping patches and reduces the number of redundant predictions.

\begin{table}[tb]
  \centering \resizebox{.93\linewidth}{!}{%
  \begin{tabular}{|c|c|c|c|c|}
    \hline
    \small \textbf{SDD} & $\gamma_1 = 1$ & $\gamma_1 = 3$  & $\gamma_1 = 5$  & $\gamma_1 = 10$ \\
    \hline 
    \footnotesize \ott   512-256                           & \textbf{0.245} & \textbf{0.361} &  \textbf{0.415}  &  \textbf{0.457}\\
    \footnotesize no groups    & 0.225          & 0.321          &  0.380           &  0.438\\
    \footnotesize fixed  offsets    & 0.199          & 0.136          &  0.246           &  0.244\\
    \footnotesize no offsets    & 0.127          & 0.127          &  0.125           &  0.122\\
    \hline
    \hline
    \footnotesize \ott   256-128                           & \textbf{0.128}         & \textbf{0.243} &  \textbf{0.293}  &  \textbf{0.331}\\
    \footnotesize no groups   & 0.122                  & 0.229          &  0.282           &  0.326\\
    \footnotesize fixed offsets   & 0.088                  & 0.136          &  0.150           &  0.154\\
    \footnotesize no offsets     & 0.030                  & 0.040          &  0.040           &  0.040\\
    \hline
  \end{tabular}
}
  \caption{\label{tab:singles}\map{@0.5} results comparing \method{} with three ablation variants, \emph{no groups}, \emph{fixed offsets} and \emph{no offsets} (see text).}
\end{table}

\subsection{Ablation study}
In this section we briefly report on ablation experiments that
 highlight the influence of the proposed contributions.
Detailed results are provided in the supplemental material.

\medskip\noindent\textbf{Memory requirements.} \method{} stages are applied consequently, hence only one network needs to live in memory at a time. 
However having independent networks for each stage can still be prohibitory when working with very large backbones, hence we also study a variant of \method{} where weights are shared across stages. While this reduces the number of model parameters, we find that it can significantly hurt detection accuracy in our settings. 
A likely explanation is that the data distribution in stage 1 and stage 2 are drastically different in terms of object resolution and distribution, effectively causing a \textit{domain shift}. 

%
\medskip\noindent\textbf{Groups.} We compare \method with a variant without group 
information: we drop the loss term ${\mathcal L}_{\text{groups}}$ 
in \hyperref[odgiloss]{(\ref{odgiloss})} and ignore group flags 
in the transition between stages.
\hyperref[tab:singles]{Table \ref{tab:singles}} (row \emph{no groups}) 
shows that this variant is never as good as \method{}, even for larger number of crops, confirming that 
the idea of grouped detections provides a consistent advantage. 

\medskip\noindent\textbf{Offsets.} 
We perform two ablation experiments to analyze the influence of 
the region rescaling step introduced in 
\hyperref[sec:offsets]{Section \ref{sec:offsets}}. 
First, instead of using learned offsets we test the 
model with offset values fixed to $\frac23$, \ie 50\% expansion of the bounding boxes, which corresponds to the value of the target offsets margin $\delta$ we chose for standard \method{}.
Our experiments in \hyperref[tab:singles]{Table \ref{tab:singles}} 
show that this variant is inferior to \method{}, confirming 
that the model benefits from learning offsets 
tailored to its predictions. 
Second, we entirely ignore the rescaling step during 
the patch extraction step (row \emph{no offsets}).
This affects the \map{} even more negatively: 
extracted crops are generally localized close to the 
relevant objects, but do not fully enclose  
them. 
Consequently, the second stage retrieves partial objects, 
but with very high confidence, resulting in strong false 
positives predictions. 
In this case, most correct detections emerge from stage 1's 
early-exit predictions, hence increasing $\gamma_1$, \ie passing forward 
more crops, does not improve the \map{} in this scenario.


\section{Conclusions}
We introduce \method{}, a novel cascaded scheme for object detection 
that identifies \emph{groups of objects} in 
early stages, and refines them in later stages \emph{as needed}: 
Consequently, (i) empty image regions are discarded, thus saving 
computations especially in situations with heterogeneous object 
density, such as aerial imagery, and (ii) groups are typically larger 
structures than individuals and easier to detect at lower resolutions.
Furthermore, \method{} can be easily added to off-the-shelf backbone 
networks commonly used for single-shot object detection:  
In extensive experiments, we show that the proposed method offers substantial 
computational savings without sacrificing accuracy. 
The effect is particularly striking on devices with limited 
computational or energy resources, such as embedded platforms.

{\small
\bibliographystyle{ieee}
\bibliography{egbib}
}

\end{document}